# Reverse Engineering Chemical Reaction Networks from Time Series Data

*Dominic P. Searson[1], Mark J. Willis[2]\* and Allen Wright[2]*

1 Northern Institute for Cancer Research, Newcastle University, United Kingdom.

2 School of Chemical Engineering and Advanced Materials, Newcastle University, United Kingdom.

d.p.searson@ncl.ac.uk; mark.willis@ncl.ac.uk; a.r.wright@ncl.ac.uk

\*Corresponding author. Tel. +44 (0)191 2227242

**Abstract**

The automated inference of physically interpretable (bio)chemical reaction network models from measured experimental data is a challenging problem whose solution has significant commercial and academic ramifications. It is demonstrated, using simulations, how sets of elementary reactions comprising chemical reaction networks, as well as their rate coefficients, may be accurately recovered from non-equilibrium time series concentration data, such as that obtained from laboratory scale reactors. A variant of an evolutionary algorithm called differential evolution in conjunction with least squares techniques is used to search the space of reaction networks in order to infer both the reaction network topology and its rate parameters. Properties of the stoichiometric matrices of trial networks are used to bias the search towards physically realisable solutions. No other information, such as chemical characterisation of the reactive species is required, although where available it may be used to improve the search process.

# 1 Introduction

Effective tools for the reverse engineering of chemical reaction networks from laboratory scale time course concentration data are likely to become of increasing commercial and academic importance. For instance, a model of a reaction network – written as a coupled set of ordinary differential equations (ODEs) describing the dynamic behaviour of the system - comprises the central numerical description of the reaction network in modern process simulation and optimisation software. Software of this nature is required for numerous reasons including accurate and economic plant design and process optimisation [1] and so methods, tools and procedures for rapidly establishing the reaction pathway from data using as little a priori information as possible are desirable.

In particular, methods that can be applied to data obtained from reaction systems operating far away from chemical or biochemical equilibrium are of interest. This is because batch and semi-batch reactors – rather than continuous stirred tank reactor (CSTRs) – tend to be used in the fine chemical and pharmaceutical industries during the chemical development lifecycle. Furthermore, the increased uptake of high throughput technologies e.g. automated robotic workstations for performing many experiments in parallel, coupled with improved *in situ* sensor technology to provide rich data sets, is likely to provide an increase in the quantity and quality of non-equilibrium experimental reaction data. Currently, this data is generally only used for kinetic fitting - once a kinetic model structure has been postulated - but there exists the potential to extract more useful information, such as reaction network topologies as well as reaction stoichiometries, reaction rates, reaction fluxes and kinetic models ([1], [2], [3]).

Without suitable prior knowledge, however, the deduction of a fully characterised, interpretable dynamic mathematical description of reaction network from observed time composition data alone is an ill-posed problem. This type of problem – which may be thought of as "reverse engineering" a useful description of a network from data – is usually very difficult to solve uniquely. For instance, when there is little a priori knowledge of the nature of the entities within the network and when the measured data is corrupted with measurement noise, there are frequently many plausible networks that can explain the data ([4],[5]).

One of the earliest advances in reaction network inference from non-equilibrium data was made by Bonvin and Rippin [2] who proposed a methodology - called target factor analysis (TFA) - to identify both the number of linearly independent reactions and also to test whether proposed reaction stoichiometries are consistent with the measured time series concentration data. The principal advantage of TFA is that no knowledge of reaction kinetics or fluxes is required. Once a set of plausible reaction stoichiometries is extracted, the user can then propose suitable ODE kinetic model descriptions and test these against the data. This work has subsequently been developed by a number of workers (e.g. [3], [6], [7], [8])

A contrasting approach is to parameterise a suitable ODE model structure directly from the observed data and to use the resulting model to infer the network properties. One such approach – referred to as the S-system methodology ([4], [5], [9], [10], [11], [12]) – has been investigated extensively in the biosciences for modelling non-equilibrium time series data (e.g. gene expression microarray data) – and forms part of the biochemical systems theoretic (BST) framework for the dynamic modelling and analysis of biological systems. The S-system model structure possesses a number of desirable properties. For instance, experimental time series can be used to estimate the parameters of S-system models that can both approximate the temporal dynamics of a connected network of entities and be used to infer the connective structure of the network. However, an estimated S-system model does not, in itself, provide any direct information about the number or nature of the reaction steps, e.g. stoichiometries or reaction orders.

Domain dependent knowledge can be exploited to narrow the network search space by restricting the form of the ODE model used to explain the dynamic behaviour of the chemical reaction network. For instance, elementary chemical and biochemical reactions, occurring in well mixed, relatively dilute, homogeneous phases – such as may be found in controlled laboratory batch and fed-batch experiments – often obey the law of mass action kinetics. This allows a class of physically interpretable ODE models with pseudo-linear properties to be formulated ([13], [14], [15], [16]). The pseudo-linear properties of these models allow classical regression techniques to be applied within the ODE model search process.

In this paper we describe an approach, closely related to the pseudo-linear approaches described above, that uses a variant of the evolutionary computational method of differential evolution (DE) to search the space of elementary chemical reaction networks. DE [17] is employed to iteratively evolve a population of numerical vectors, each of which encodes the topology of a trial network of elementary reactions. Each trial network has a corresponding stoichiometric matrix, the properties of which may be used to determine whether the network is physically realisable. The objective function for the DE is designed so that the search is biased towards networks that are physically possible. In addition, for each trial network, the rate coefficients – and hence the vector of reaction rates - are identified from the estimated time derivatives of the measured concentration data using multiple linear regression. It is shown, using realistic amounts of noise corrupted time series data acquired from simulations of several chemical reaction networks in batch experiments, that the approach is effective at recovering the "true" network of elementary reactions and providing good estimates of the associated rate coefficients.

The paper is structured in the following way. In Section 2, mathematical models of elementary reaction networks are discussed in order to provide a definition of the particular network inference problem. In Section 3, the numerical representation of possible network topologies is described. In Section 4, the objective function used for network search is presented. The procedure used to determine whether trial networks are physically valid is also described for cases where chemical information is either available or not. In Section 5, the basic DE method and the self-adaptive variant of DE used to search for reaction networks is described. In Section 6, three case studies using simulated chemical reaction systems are presented. Finally, in Section 7 some conclusions and a discussion are presented.

## 2 Problem definition

A materials balance for $S$ chemical components taking place in $R$ chemical reactions comprising a chemical reaction network can be written for certain classes of chemical reactors – batch, fed-batch (semi-batch) and continuous flow stirred tank reactor (CSTRs). These balances are often written to reflect the pragmatic assumptions that the (homogeneous phase) reactor is operating isothermally, is well mixed and that the overall density of the reaction mixture is not significantly changed by the occurrence of the chemical reactions within the reactor. In this paper we will, without loss of generality, consider batch reactions only, i.e. there are no materials added or removed during the course of an experiment. In this case a materials balance for each species can be written as:

$$\frac{d[x_i]}{dt} = f_i \quad i = 1,\ldots,S \tag{1}$$

where $[x_i]$ is the molar concentration of species $i$ at time $t$. Eqn. (1) is a set of coupled ODEs that describes the dynamic behaviour of the reactive species due to chemical reactions, as represented by the $S$ reaction fluxes $f_i$.

The flux terms $f_i$ are directly linked to the stoichiometries of the $R$ reactions taking place and the kinetic rate terms of these reactions. For instance, consider the $(S \times R)$ stoichiometric matrix **N** containing the stoichiometric coefficients for each of the $S$ species in the $R$ reactions.

$$\mathbf{N} = \begin{bmatrix} n_{11} & \cdots & n_{1R} \\ \cdots & \ddots & \cdots \\ n_{S1} & \cdots & n_{SR} \end{bmatrix} \qquad (2)$$

Here, $n_{ij}$ is the stoichiometric coefficient of the $i$th chemical species in the $j$th reaction. By convention, $n_{ij} < 0$ for a species that undergoes net consumption in a reaction, $n_{ij} > 0$ for a species that undergoes net production and $n_{ij} = 0$ for a species that is either not involved in the $j$th reaction or has no net change in it. The component balances may now be conveniently expressed in terms of $\mathbf{N}$ and the $R$ individual reaction rates in matrix-vector form:

$$\frac{d\mathbf{x}}{dt} = \mathbf{Nr} \qquad (3)$$

where $\mathbf{r}$ is the $(R \times 1)$ dimensional vector of reaction rates and $\mathbf{x}$ is the $(S \times 1)$ dimensional vector of species concentrations $[x_1],\ldots,[x_S]$ at time $t$. The $R$ reaction rates constituting $\mathbf{r}$ are, in general, non linear functions of the concentrations $\mathbf{x}$ and linear functions of the rate coefficients. If elementary reactions are assumed then the form of the $R$ rate terms in $\mathbf{r}$ is determined uniquely by the reactants in each of the $R$ elementary equations.

The law of mass action kinetics states that the rate of an elementary reaction may be assumed to be directly proportional to the collision frequency of the reactants, and hence the product of the reactant concentrations. Consider a bimolecular elementary chemical reaction with rate coefficient $k$ involving $a$ molecules of species $x_1$ and $b$ molecules of species $x_2$ forming $c$ molecules of species $x_3$ and $d$ molecules of species $x_4$.

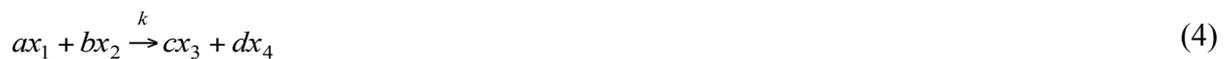

$$ax_1 + bx_2 \xrightarrow{k} cx_3 + dx_4 \qquad (4)$$

The rate $r$ of this reaction at any time, according to the law of mass action, is:

$$r = k[x_1]^a [x_2]^b \qquad (5)$$

For example, consider the hypothetical chemical reaction network comprising four reactive species $x_1,\ldots, x_4$ involved in two elementary reactions with rate coefficients $k_1$ and $k_2$:

$$x_1 + x_2 \xrightarrow{k_1} x_3 + x_4 \quad x_3 \xrightarrow{k_2} x_4 \tag{6}$$

The stoichiometric matrix **N** and rate vector **r** for this network are:

$$\mathbf{N}^T = \begin{bmatrix} -1 & -1 & 1 & 1 \\ 0 & 0 & -1 & 1 \end{bmatrix} \quad \mathbf{r} = \begin{bmatrix} k_1[x_1][x_2] \\ k_2[x_3] \end{bmatrix} \tag{7}$$

Hence the $S$ ODEs, for known initial conditions of the $[x_i]$, can be used to describe the temporal evolution of the species concentrations:

$$\frac{d\mathbf{x}}{dt} = \mathbf{Nr} = \begin{bmatrix} -k_1[x_1][x_2] \\ -k_1[x_1][x_2] \\ k_1[x_1][x_2] - k_2[x_3] \\ k_1[x_1][x_2] + k_2[x_3] \end{bmatrix} \tag{8}$$

Note that any given row of **N** does not uniquely correspond to an elementary reaction, nor does any **N** correspond to a unique set of elementary reactions. For instance, the following reaction network has the same **N** as the network described by Eqn. (6).

$$x_1 + x_2 \xrightarrow{k_1} x_3 + x_4 \quad x_1 + x_3 \xrightarrow{k_2} x_1 + x_4 \tag{9}$$

Using the development above, the general problem of inferring a network of elementary reactions from time series concentration data can be stated as: using only noise corrupted measurements of **x** sampled from $M$ experiments reconstruct one or more plausible elementary reaction networks, i.e. the number and nature of the elementary reactions, the corresponding stoichiometric matrix **N**, the rate coefficients and the reaction rates **r**.

## 3 Reconstruction of elementary reaction networks from data by network search

Recently, Srividhya *et al.* [15] described a method to reconstruct (bio)chemical reaction networks from time series concentration data method by searching the space of networks of elementary unimolecular and bimolecular reactions in order to minimise an objective function that takes into account both the predictive ability and the parsimony of the proposed reaction network. This search is stepwise in nature and is accomplished by adding or removing one reaction at a time from previously high performing networks of elementary unimolecular and bimolecular reactions. For each network (i.e. set of elementary reactions) evaluated, the corresponding reaction rate terms, multiplied by the appropriate stoichiometric coefficients, are used to set up a constrained regression problem in which the network rate coefficients are computed from the estimated derivatives of the time concentration data using a non-negative least squares method. The minimisation of an objective function that trades off the mean squared prediction error of the estimated derivatives against the number of reactions in each trial set is used to drive the stepwise search process. The advantages of this approach are that it is automated, non-subjective and that it exploits the fact that the rates of change of concentration (the fluxes $f_i$) of the $S$ species are linear combinations of the same $R$ reaction rate terms. The principal disadvantages of this approach as presented in [15] are that firstly, it uses a stepwise method to search the space of reaction networks. Stepwise methods are essentially local search operators and are likely to be trapped in local minima, especially in noisy, high dimensional, multi-modal search spaces such as that typically encountered in network inference. This appears to be borne out by some the results presented in [15] in which the stepwise approach fails to recover the correct network for some fairly low complexity bimolecular reaction networks under noise-free conditions. Secondly, the search method does not penalise physically unrealisable reaction networks. For instance, consider the following trial network:

$$2x_1 \xrightarrow{k_1} x_2 \qquad x_1 \xrightarrow{k_2} x_2 \tag{10}$$

This is not physically possible, but such a network could nominally satisfy an objective function based on prediction error and network complexity alone. Thirdly, the method makes no provision for the use of a priori information about the reactive species (e.g. molecular weights, chemical composition).

In an attempt to address these issues, a method which takes into account these factors is presented in this paper.

### 3.1 Network search as a non-linear integer programming problem

In order to search through the space of reaction networks it is first necessary to employ a suitable numerical representation of possible network topologies. In this paper, the topology of a reaction network containing up to $q_{max}$ reactions is represented by a vector of integers. The search of the network space may then be formulated as the search for the optimal integer vector, i.e. an integer programming problem. An optimal solution is that which best satisfies an objective function that balances network complexity and physical validity against how well the network fits the data.

Our integer representation of each of the $q_{max}$ reactions is simply a description of the identities of the reactants and the products in each reaction. For instance, consider the case where the reactions are restricted to be either unimolecular or bimolecular with a maximum of two reactants and three products[i]. Each reaction can then be represented by a sequence of five integers. The first two integers represent the identities of the reactants and the remaining three integers represent the identities of the products. A zero is used to "switch off" a reactant or product in a reaction. The purpose of this is to allow the representation of reactions of lower complexity than the specified maximum. Hence, each entry in the integer sequence can take a value in the range 0 to $S$.

The topology of the whole network is represented by concatenating the $q_{max}$ integer sequences into one vector. This is illustrated in Figure 1 where $q_{max} = 4$ and $S = 3$. The reaction representation rule is also illustrated for the five integers representing reaction 1.

---

[i] The integer representation used can be easily modified to encompass reaction types of greater or lesser complexity than that illustrated.

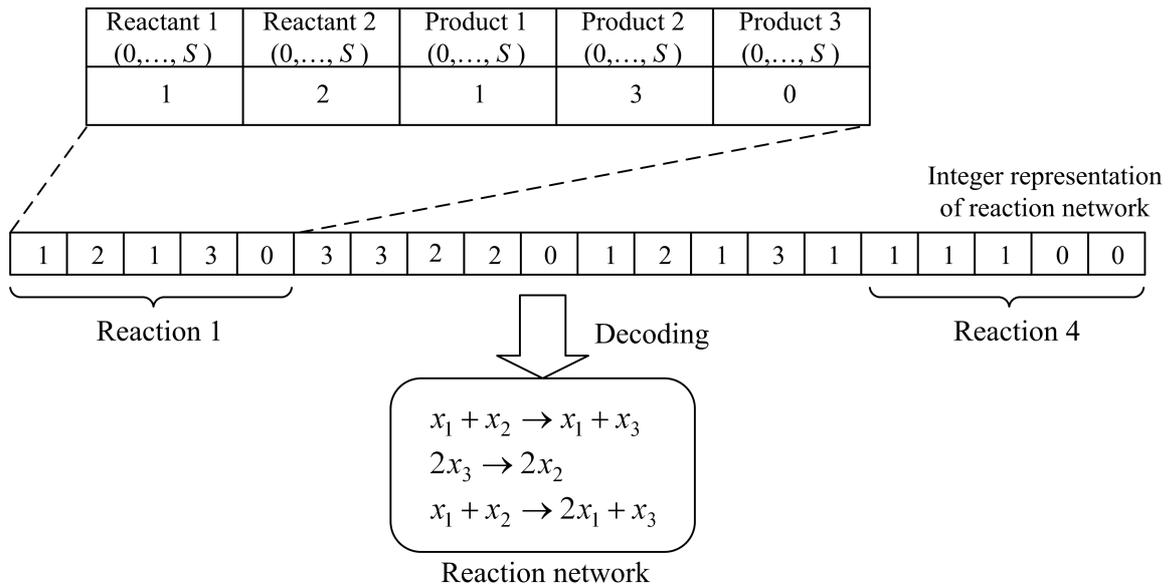

**Figure 1**. Example of integer vector representation of the topology of a trial reaction network containing $S = 3$ species and a maximum of $q_{max} = 4$ reactions. In this case, reaction 4 is invalid and is excluded from the overall network.

Note that the representation also allows the specification of physically unrealisable reactions, such as reaction 4 in Figure 1 which is:

$$2x_1 \rightarrow x_1 \qquad (11)$$

Invalid reactions such as these are interpreted as "no reaction occurring" and are excluded from the trial network. Hence the actual size $q$ of the network in Figure 1 is 3. This is accomplished by examining the columns of the trial network's stoichiometric matrix and removing those that do not contain at least one positive and one negative coefficient. The ability to "switch off" reactions is an important feature because it is not assumed exactly how many reactions $R$ are occurring in the unknown reaction network, but it is assumed that $R < q_{max}$ where $q_{max}$ is set to some "sensible" user defined limit.

Whilst this representation excludes individual reactions that are physically unrealisable it does not prohibit entire networks with this property. The trial network illustrated in Figure 1, for instance, is not physically possible because the third reaction is inconsistent with the first and second reactions. Such networks are not rejected out of hand, however, because any integer programming method employed to search the space of network topologies would require substantial modification in order to search the integer vector space in such a way that only trial solutions corresponding to physically realisable networks are generated. This would then limit the applicability and simplicity of the approach. Instead, we employ a soft constraint to guide the DE based integer programming method towards physically realisable solutions. This is discussed further in Section 4.

### 3.2 Estimation of the rate coefficients for trial reaction networks

To evaluate how well any trial network topology fits the data it is necessary to estimate the rate coefficients. However, parameter estimation for each of the corresponding ODE models from time series concentration data usually requires the use of iterative optimisation techniques (e.g. gradient descent methods). This typically involves repeated numerical solution of the ODEs using many trial parameter sets until the simulation closely matches the experimental data (i.e. "kinetic fitting"). A drawback of this approach is that the numerical solution of the ODEs for each trial parameter set for each of the evaluated trial network topologies is - currently - prohibitively computationally expensive.

To avoid the computational expense of numerical integration it is possible – prior to the network search procedure - to estimate the time derivatives for each of the $S$ species from the measured time series concentration data [18]. This can be achieved by fitting non-linear "smoothing" functions to the concentration measurements and either numerically or analytically obtaining the derivatives at the required sample times from the fitted function. Examples of this in the literature include the use of forward differencing [15], feedforward artificial neural networks [18] and rational polynomials [16].

Once estimates of the time derivatives are obtained, the rate coefficients for each trial network can be estimated from the data using least squares methods. This requires the formation of a response vector **y** and a predictor (model design) matrix **X** for each trial network. These can be related to the ($q \times 1$) vector of the unknown rate coefficients **k** by a linear equation that may be solved using least squares to minimise the vector of errors **ε**.

$$\mathbf{y} = \mathbf{Xk} + \boldsymbol{\varepsilon} \tag{12}$$

Following the development in [15], this is outlined below for a single experiment. Let $[x_i]_n$ represent the sampled concentration of the $i$th species at the $n$th time point during an experiment, where there are $N$ time points, and let $D_{i,n}$ represent the estimated time derivative of the $i$th species concentration at the $n$th time point. An ($N \times q$) matrix $X_i$ can be formed for each species, the $j$th column of which comprises the unscaled kinetic rate term for the $j$th reaction, evaluated at each time point, multiplied by a coefficient $v_{i,j}$. This coefficient is the molecularity of the $i$th species in the $j$th reaction multiplied by +1 if the the $i$th species is a product in the $j$th reaction, by -1 if it is a reactant, and by 0 if it does not take part in the $j$th reaction. For example, in the following trial network the coefficients are $v_{1,1} = -2$, $v_{1,2} = 0$, $v_{2,1} = 1$, $v_{2,2} = -1$, $v_{3,1} = 0$ and $v_{3,2} = 1$.

$$2x_1 \xrightarrow{k_1} x_2 \quad x_2 \xrightarrow{k_2} x_3 \tag{13}$$

The $X_i$ matrices for this network are shown in Eqn. (14).

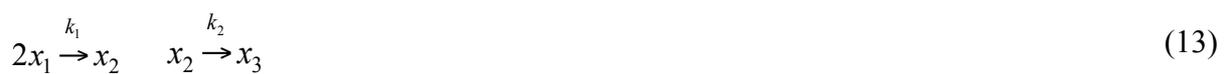

$$\mathbf{X}_1 = \begin{bmatrix} -2[x_1]^2_{n=1} & 0 \\ -2[x_1]^2_{n=2} & 0 \\ \vdots & \vdots \\ -2[x_1]^2_{n=N} & 0 \end{bmatrix} \quad \mathbf{X}_2 = \begin{bmatrix} [x_1]^2_{n=1} & -[x_2]_{n=1} \\ [x_1]^2_{n=2} & -[x_2]_{n=2} \\ \vdots & \vdots \\ [x_1]^2_{n=N} & -[x_2]_{n=N} \end{bmatrix} \quad \mathbf{X}_3 = \begin{bmatrix} 0 & [x_2]_{n=1} \\ 0 & [x_2]_{n=2} \\ \vdots & \vdots \\ 0 & [x_2]_{n=N} \end{bmatrix} \tag{14}$$

The $X_i$ are then vertically concatenated to form the ($SN \times q$) overall predictor matrix **X**. Similarly, the ($SN \times 1$) response vector **y** is formed from the vertical concatenation of the $S$ vectors $y_i$ each of which contains the $N$ estimated time derivatives $D_{i,n}$. The extension to $M$ multiple experiments is straightforward and yields an **X** of dimension ($MSN \times q$) and **y** of dimension ($MSN \times 1$).

To compute **k** we use the least squares normal equation with the Moore-Penrose pseudo-inverse $(\mathbf{X}^T\mathbf{X})^{\#}$ to mitigate problems with co-linearity in **X**:

$$\mathbf{k} = (\mathbf{X}^T\mathbf{X})^{\#}\mathbf{X}^T\mathbf{y} \tag{15}$$

In our method, any reactions that yield rate coefficients that are negative or zero are simply deleted from the trial network. The corresponding elements in **k** and columns in **X** are also then deleted. The sum of squares of prediction errors of the estimated derivatives ($SSE_D$) for the rectified trial network can then be calculated as:

$$SSE_D = (\mathbf{y} - \mathbf{Xk})^T(\mathbf{y} - \mathbf{Xk}) \tag{16}$$

In the degenerate case, where there are no reactions remaining in the rectified trial network ($q = 0$), a value of infinity is assigned to the $SSE_D$.

**4 Formulation of the objective function for network search**

An objective function based purely on the minimisation of prediction errors does not take into account the desired model characteristics. It is preferable to use one of the formal statistical metrics (information criteria; IC) that can be minimised to in order to balance the trade-off between model complexity (in this case, the number of reactions $q$) and how well the model fits the data, e.g. Akaike's information criterion (AIC; [19]). However, it is seldom clear which is the "best" criterion to use. A number of ICs were tested and it was found that the Schwarz information criterion (SIC; [20]) appeared to yield the best all round performance. The SIC – sometimes referred to as the Bayesian information criterion - for a trial network can be written as:

$$\text{SIC}_D = N_s \ln\left(\frac{\text{SSE}_D}{N_s}\right) + p\ln(N_s) \qquad (17)$$

where $N_s$ is the sample size and $p = q + 1$ [21].

The SIC measure, however, does not discriminate between physically valid and physically unrealisable reaction networks. To guide the network search process, we propose the use of an additional term in the objective function to penalise trial networks that violate physical constraints. This is achieved my modifying the objective function in Eqn. (17) as follows:

$$\text{SIC}_D = N_s \ln\left(\frac{\text{SSE}_D}{N_s}\right) + (p+\alpha)\ln(N_s) \qquad (18)$$

where $\alpha = 0$ for physically valid networks and $\alpha = q_{max}$ for physically invalid networks. The meaning of this is that physically invalid networks are penalised in terms of the SIC measure as if they were comprised of $q + q_{max}$ elementary reactions rather than $q$ reactions. Hence, physically unrealisable trial networks are not considered infeasible solutions - but they are strongly penalised. In terms of the evolutionary search process, this allows relatively high performing partial solutions – although formally invalid - to be explored early on in the expectation that these will be combined with other high performing partial solutions later in the search process. As the search progresses, the evolutionary process increasingly discriminates in favour of relatively high performing valid solutions (i.e. those for which $\alpha = 0$). In this case the modified objective function reverts to the standard SIC.

In order to determine whether a trial network is physically valid, two scenarios are considered. The first scenario is where additional physical information and/or chemical characterisation of the reactive species is available. The second scenario is where no additional physical information is available. These will be considered in turn.

### 4.1 Physical/chemical information available

Physical and chemical properties of the reaction network that are conserved are referred to as reaction invariant properties. For example, the molecular weight, number of carbon atoms and number of oxygen atoms of a particular species do not change regardless of the nature and number of the chemical reactions occurring. If $C$ such properties are known for the $S$ species then a $(C \times S)$ conservation matrix $\mathbf{A}$ may be formed, where the entry at the $i$th column and $j$th row refers to the value of the $j$th conserved property for the $i$th species. The simplest case of $\mathbf{A}$ is probably the $(1 \times S)$ vector containing the molecular weights – or the ratios of the molecular weights - of the $S$ species. Due to the conservation of the properties within $\mathbf{A}$, the following relationship holds for physically realisable reaction networks [22].

$$\mathbf{AN} = 0 \tag{19}$$

where $\mathbf{N}$ is the $(S \times q)$ stoichiometric matrix corresponding to the reaction network. Note that this property is independent of the kinetic behaviour of the network. If the relationship in Eqn. (19) does not hold for a trial network in the network search process, then α is set to $q_{max}$ in the search objective function, as indicated in Section 3.

### 4.2 No physical/chemical information available

If no conserved physical or chemical properties are available for the $S$ species then a conservation matrix $\mathbf{A}$ can not be explicitly formed. However, it is still possible to make use of the relation given in Eqn. (19) even if $\mathbf{A}$ is unknown. This is because the transpose of any valid conservation matrix $\mathbf{A}^T$ resides in the nullspace of the transpose of the trial network's stoichiometric matrix $\mathbf{N}^T$. This can be seen by rewriting Eqn. (19) as:

$$\mathbf{N}^T\mathbf{A}^T = 0 \tag{20}$$

Therefore, if $\mathbf{N}^T$ does not have a nullspace then no conserved physical or chemical properties matrix $\mathbf{A}$ can exist and thus the network is physically unrealisable. The nullspace of $\mathbf{N}^T$ for any trial network can be calculated by means of the singular value decomposition. In practice, commands for computing nullspaces of matrices are built in to most modern numerical computing software, e.g. Mathematica, MATLAB etc. Usually the nullspace is computed in the form of a set of orthonormal basis vectors.

There are also trial networks for which the nullspace of $\mathbf{N}^T$ does exist but has rows that contain only zeros. In these cases, a solution of Eqn. (20) is algebraically possible, but the solution does not correspond to a set of physically possible conserved properties. For instance, the following reaction network topology has a nullspace of $[-0.7071\ -0.7071\ 0]^T$ :

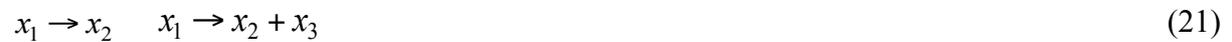

$$x_1 \rightarrow x_2 \qquad x_1 \rightarrow x_2 + x_3 \tag{21}$$

The interpretation of this is that $\mathbf{A}$ exists in the algebraic sense but that it only does so if *all possible* conserved properties for $x_3$ are zero. However, the set of all possible conserved properties includes properties that cannot have zero values, e.g. mass. Therefore, such solutions are not physically realisable and so the trial network is penalised in Eqn. (18).

In summary, if the nullspace of $\mathbf{N}^T$ for a trial network either does not exist or contains one or more rows that contain only zeros, then α is set to $q_{max}$ in the network search objective function.

The network topology representation, rate coefficient estimation and the objective function for the network search problem have been defined in this and previous sections. In the next section, the population based evolutionary method of differential evolution for searching through the space of networks is discussed.

## 5 Differential evolution for searching the space of reaction networks

Evolutionary algorithms (EAs) are inherently well suited for many search and optimisation problems: they can find near optima in astronomically large, multi-modal search spaces and they are robust across many problem classes. Differential evolution (DE) is an EA that was originally conceived as a continuous variable optimisation method [17], and although not well very known, studies have shown DE to outperform other EAs for a variety of benchmark problems [23]. Further advantages of DE are that (a) it is compact and very simple to implement (b) it has relatively few parameters to set (c) it may easily be modified to operate with discrete, integer and continuous variables and mixtures thereof and (d) it can operate both as unconstrained optimiser and as a constrained optimiser with multiple constraints [24]. Before discussing the variant of DE used here, the "basic" DE method for continuous optimisation is outlined[ii].

### 5.1 Basic DE optimisation method

Consider a problem involving finding a vector **z** comprising the $n$ continuous parameters $z_1$, …, $z_n$ with values that minimise an objective function of these parameters $f(\mathbf{z})$. The parameters in **z** generally have upper and lower bounds defined individually by the user for each element of **z**. In order to minimise $f(\mathbf{z})$, DE performs search by simulated evolution on a population of $P$ such trial vectors $\mathbf{z}_{i,g}$ over $G$ generations where $i = 1,…, P$ and $g = 1, …, G$. The $i$th vector in the $g$th population contains the $n$ individual trial parameters $z_{i,j,g}$ where $j = 1, …, n$.

The underlying mechanism of DE is that each vector $\mathbf{z}_{i,g}$ in the current population is used to generate a single mutated offspring $\mathbf{m}_{i,g}$. The mutated offspring is then compared directly with its parent on the objective function. If the offspring equals or surpasses its parent it replaces it in the next generation, otherwise the parent is retained. Eventually, the evolutionary processes of mutation and selection should drive the population towards the optimum values of the parameter vector **z**. The steps in the DE algorithm are outlined below.

---

[ii] In the optimisation literature this version of DE is referred to as DE/rand/1/bin.

Step 1. DE initialisation

The algorithm is initialised in the first generation $g = 1$ by randomly generating the $P$ vectors $\mathbf{z}_{i,1}$ within the ranges specified by the upper and lower bounds for each parameter. Each $\mathbf{z}_{i,1}$ is then evaluated against the objective function in order to compute the values $f(\mathbf{z}_{i,1})$.

Step 2. DE Mutation

The mutation vector $\mathbf{m}_{i,g}$ is created by replacing a probabilistically determined number of the elements of $\mathbf{z}_{i,g}$ with corresponding elements from an "exploration" or "perturbation" vector $\mathbf{v}_{i,g}$. The perturbation vector itself is created by adding the scaled difference of two randomly selected vectors $\mathbf{z}_{a2,g}$ and $\mathbf{z}_{a3,g}$ from the current population to another randomly selected vector $\mathbf{z}_{a1,g}$ from the current population where $i \neq a1 \neq a2 \neq a3$.

$$\mathbf{v}_{i,g} = \mathbf{z}_{a1,g} + F(\mathbf{z}_{a2,g} - \mathbf{z}_{a3,g}) \tag{22}$$

The "scaling factor" $F$ is a user defined parameter of the DE and is often set in the range $0 < F \leq 2$. The other user defined parameters are the population size $P$, the number of generations $G$ and the "crossover rate" $CR$. The latter falls in the range $0 \leq CR \leq 1$ and is used to probabilistically determine what proportion of the elements of the current member $\mathbf{z}_{i,g}$ are replaced with elements of the perturbation vector $\mathbf{v}_{i,g}$ when creating $\mathbf{m}_{i,g}$.

The mutated offspring vector $\mathbf{m}_{i,g}$ is formed by means of a binomial crossover mechanism on $\mathbf{z}_{i,g}$ and $\mathbf{v}_{i,g}$ as follows. To determine the $j$th element $m_{i,j,g}$ in $\mathbf{m}_{i,g}$ a random number $a_j$ is generated in the range $0 \leq a_j \leq 1$. If $a_j < CR$ then $m_{i,j,g} = v_{i,j,g}$ otherwise $m_{i,j,g} = z_{i,j,g}$. That is, if the random number is below the crossover rate then a parameter value from the perturbation vector is used and, if not, the parameter value from the current individual is retained. The values in $\mathbf{m}_{i,g}$ are then checked against the upper and lower bound constraints for each parameter. If any element of $\mathbf{m}_{i,g}$ violates the bounds then it can either be replaced with a random number generated within the bounds or truncated to the exceeded bound value. In this work, the latter option is used.

Step 3. DE competition and population update

Each mutated vector is then evaluated against the objective function $f(\mathbf{m}_{i,g})$ and the resulting value compared to that obtained by its parent $f(\mathbf{z}_{i,g})$. If $f(\mathbf{m}_{i,g}) \leq f(\mathbf{z}_{i,g})$ then $\mathbf{m}_{i,g}$ takes the place of $\mathbf{z}_{i,g}$ in the next generation and $\mathbf{z}_{i,g+1} = \mathbf{m}_{i,g}$. Otherwise, the current solution $\mathbf{z}_{i,g}$ is retained and copied to the next generation and $\mathbf{z}_{i,g+1} = \mathbf{z}_i$. The generation counter $g$ is then set to $g+1$ and the algorithm returns to Step 2.

Step 4. DE termination

The above process is repeated until some termination criterion is reached, most commonly after the fixed number of generations $G$ has elapsed. At this point the vector $\mathbf{z}_{i,G}$ in the final population with the lowest value of the objective function is usually selected as the "best result" of the run. This does not have to be the case, however, as will be seen in Section 6.

**5.2 Self-adaptive DE with integer variables**

We use a self-adaptive variant of DE where the evolved parameters vectors are interpreted as integer vectors that encode trial reaction network topologies as described in Section 3 using the modified SIC objective function in Eqn. (18). Modifying DE to produce vectors that contain integer elements is simple. The DE algorithm essentially operates as described above on continuous parameter vectors and the only difference is that, prior evaluating a vector $\mathbf{z}_{i,g}$ (or $\mathbf{m}_{i,g}$) against the objective function, the elements of $\mathbf{z}_{i,g}$ (or $\mathbf{m}_{i,g}$) are temporarily rounded to the nearest integer. Hence, by defining appropriate upper and lower bounds for each of the elements in $\mathbf{z}$ the DE can be made to operate as a non-linear integer search method within specified ranges for each parameter. Note that the underlying vector representation within the DE remains continuous and that the rounded parameter vectors are only used for the purpose of being evaluated by the objective function.

For the integer network representation described in Section 3.3 the lower and upper bounds for each continuous parameter $z_j$ in $\mathbf{z}$ are defined such that $(-1/2 + \delta) \leq z_j \leq (S + \frac{1}{2} - \delta)$ where $\delta$ is a some very small number. This is to ensure that equal lengths of the real number line are rounded correctly to the integers between 0 and $S$.

One of the drawbacks of the version of DE outlined above is that it necessary to determine "good" values of the user defined control parameters, in particular the scaling factor $F$ and crossover rate $CR$. The optimal values for these control parameters are, in general, problem dependent and usually a trial and error approach is required to tune them satisfactorily. To counter this, self-adaptive variants of DE exist in which $F$ and $CR$ are themselves evolved along with the parameters required to solve the problem. We use the self-adaptive DE proposed by Brest *et al*. [25] because it has been shown – using a suite of non-linear benchmark problems - that it equals or outperforms standard DE and other self-adapting DE variants. This self-adaptive modification to the DE method requires only minimal changes and is easy to implement. Each parameter vector in the population $\mathbf{z}_{i,g}$ is linked to corresponding values of $F$ and $CR$, referred to as $F_{i,g}$ and $CR_{i,g}$. In the initial population the values for $F_{i,g}$ and $CR_{i,g}$ are randomly generated in the ranges $0.1 \leq F_{i,g} \leq 1$ and $0 \leq CR_{i,g} \leq 1$. In Step 2 of the DE algorithm, before creating the mutant offspring $\mathbf{m}_{i,g}$, the values of $F_{i,g}$ and $CR_{i,g}$ are first mutated to new values $\underline{F}_{i,g}$ and $\underline{CR}_{i,g}$. The mutated control parameters are then used instead of fixed values of $F$ and $CR$ to create $\mathbf{m}_{i,g}$. The $\underline{F}_{i,g}$ and $\underline{CR}_{i,g}$ are created using the following probabilistic rules [25].

$$\underline{F}_{i,g} = \begin{cases} F_l + a_1 F_u & \text{if } a_2 < \tau_1 \\ F_{i,g} & \text{otherwise} \end{cases}$$

$$\underline{CR}_{i,g} = \begin{cases} a_3 & \text{if } a_4 < \tau_2 \\ CR_{i,g} & \text{otherwise} \end{cases} \quad (23)$$

where the $a_1, \ldots, a_4$ are independent random numbers generated between 0 and 1 each time the computations in Eqn. (23) are carried out and $\tau_1$ and $\tau_2$ are mutation probabilities. The rules have the effect of mutating $F_{i,g}$ to a new value between $F_l$ and $(F_l + F_u)$ with probability $\tau_1$ and mutating $CR_{i,g}$ to a new value between 0 and 1 with probability $\tau_2$. As in [25] we set $\tau_1 = \tau_2 = 0.1$ and $F_l = 0.1$ and $F_u = 0.9$.

If the mutated vector $\mathbf{m}_{i,g}$ outperforms its parent $\mathbf{z}_{i,g}$ then the values of the control parameters in the following generations are updated as $F_{i,g+1} = \underline{F}_{i,g}$ and $CR_{i,g+1} = \underline{CR}_{i,g}$. If not, the orginal values are kept and $F_{i,g+1} = F_{i,g}$ and $CR_{i,g+1} = CR_{i,}$

## 6 Network identification case studies

The proposed methods were evaluated using simulations of three hypothetical chemical reaction networks occurring in well-mixed, isothermal batch reactors within an inert solvent. The first reaction network comprises 5 species, the second 6 species and the third comprises 10 species. Reaction networks 1 and 2 are used to evaluate the methodology when there is no knowledge of any conserved physical properties and reaction network 3 is used to evaluate it when the ratios of the molecular weights of the reacting species are known.

To generate the simulated experimental data, the corresponding ODEs for each network were numerically solved using an accurate adaptive step length solver. Periodically sampled concentration values were then obtained for each species in each experiment. Arbitrary time and concentration units are used.

To simulate the effects of experimental error, the sampled concentration values for each species in each experiment were independently corrupted with additive Gaussian noise. The noise was specified to have a standard deviation equal to 4 % of the range of the concentration signal in the experiment. The reaction systems are described below:

Reaction network 1

The first hypothetical reaction network comprises five chemical species labelled $x_1, \ldots, x_5$ involved in four reactions with rate coefficients $k_1 = 0.1$, $k_2 = 0.2$, $k_3 = 0.13$ and $k_4 = 0.3$ as shown in Eqn. (24). The initial reactants are $x_1$ and $x_2$. Four experiments were simulated in which the initial concentrations were: $\{[x_1] = 0.33, [x_2] = 1\}$, $\{[x_1] = 1, [x_2] = 0.33\}$, $\{[x_1] = 1, [x_2] = 1\}$ and $\{[x_1] = 0.75, [x_2] = 1\}$. In each case, the species concentrations were sampled every time unit from 0 to 24 time units.

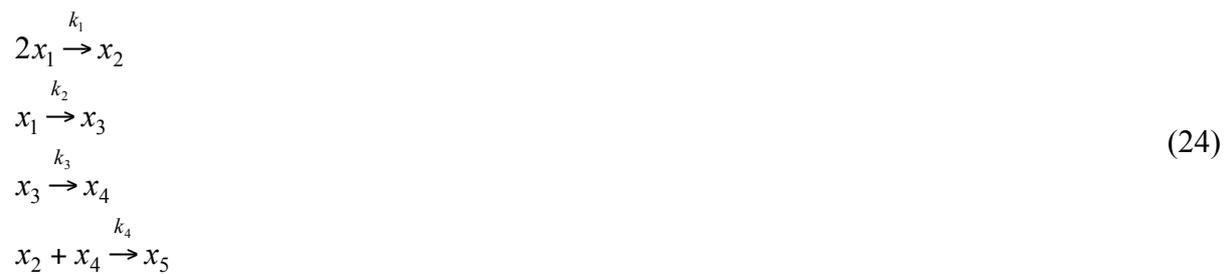

$$\begin{aligned} 2x_1 &\xrightarrow{k_1} x_2 \\ x_1 &\xrightarrow{k_2} x_3 \\ x_3 &\xrightarrow{k_3} x_4 \\ x_2 + x_4 &\xrightarrow{k_4} x_5 \end{aligned} \qquad (24)$$

Figure 2 shows the data acquired from a typical experiment. For each species the "true" unknown concentration values are plotted against the noisy sampled data.

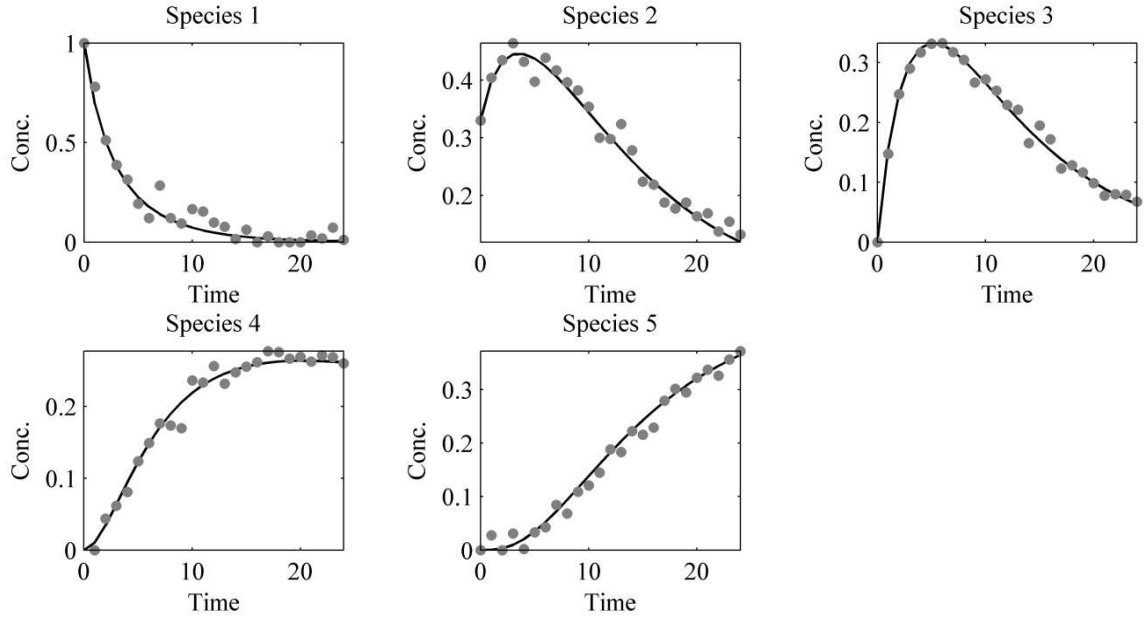

**Figure 2.** Simulated experimental concentration vs. time data for reaction network 1 (Expt. 2). The noisy sampled values (dots) are plotted against the underlying concentration values (lines).

Reaction network 2

The second reaction network comprises six chemical species labelled $x_1, \ldots, x_6$ involved in four chemical reactions with rate coefficients $k_1 = 0.2$, $k_2 = 0.1$, $k_3 = 0.15$ and $k_4 = 0.05$ as shown in Eqn. (25). Note that the fourth reaction is the reverse of the third reaction.

$$\begin{aligned} x_1 + x_2 &\xrightarrow{k_1} x_3 + x_4 \\ x_2 + x_3 &\xrightarrow{k_2} x_5 \\ x_1 + x_4 &\xrightarrow{k_3} x_6 \\ x_6 &\xrightarrow{k_4} x_1 + x_4 \end{aligned} \qquad (25)$$

The initial reactants are $x_1$ and $x_2$. Again, four experiments were simulated with the following initial concentration values: $\{[x_1] = 2.5, [x_2] = 2.5\}$, $\{[x_1] = 2.5, [x_2] = 7.5\}$, $\{[x_1] = 7.5, [x_2] = 2.5\}$ and $\{[x_1] = 10, [x_2] = 5\}$. In each case, the species concentrations were sampled every 0.5 time units from 0 to 15 time units.

Reaction network 3

The third network comprises ten chemical species labelled $x_1, \ldots, x_{10}$ involved in six chemical reactions with rate coefficients $k_1 = 0.35$, $k_2 = 0.25$, $k_3 = 0.3$, $k_4 = 0.4$, $k_5 = 0.3$ and $k_6 = 0.1$ as shown in Eqn.(26).

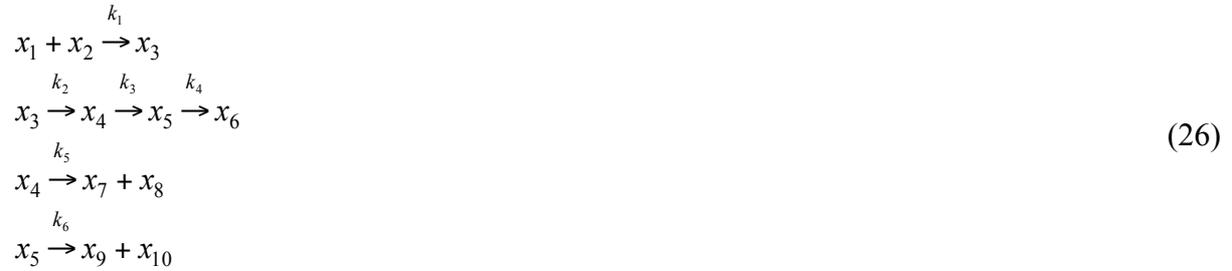

$$\begin{aligned} x_1 + x_2 &\xrightarrow{k_1} x_3 \\ x_3 &\xrightarrow{k_2} x_4 \xrightarrow{k_3} x_5 \xrightarrow{k_4} x_6 \\ x_4 &\xrightarrow{k_5} x_7 + x_8 \\ x_5 &\xrightarrow{k_6} x_9 + x_{10} \end{aligned} \quad (26)$$

It is assumed that chemical information, in the form of molecular weights, is available for each of the species in this network and so a (1 × 10) conservation matrix **A** comprising the ratios of the molecular weights, was formed as follows:

$$\mathbf{A}^T = [1 \quad 3 \quad 4 \quad 4 \quad 4 \quad 4 \quad 1 \quad 3 \quad 2 \quad 2] \quad (27)$$

Again, the initial reactants are $x_1$ and $x_2$. For this network, three experiments were simulated with the following initial concentration values: $\{[x_1] = 5, [x_2] = 5\}$, $\{[x_1] = 2, [x_2] = 4\}$, $\{[x_1] = 4, [x_2] = 2\}$. In each experiment, the species concentrations were sampled every 0.1 time units from 0 to 5 time units.

## 6.2 Estimation of time derivatives

Once the simulated experimental time series concentration data was generated, the time derivative of each species concentration value in each batch was approximated. This was accomplished by fitting a separate second order rational polynomial of the form in Eqn. (28) to each signal where $t$ is the time value and $[\hat{x}]_t$ is the predicted value of the $i$th species concentration at time $t$.

$$[\hat{x}_i]_t = \frac{p_1 t^2 + p_2 t + p_3}{t^2 + p_4 t + p_5} \qquad (28)$$

For each signal, the coefficients $p_1, \ldots, p_5$ were estimated from the noisy concentration values $[x_i]_t$ using the Levenberg-Marquardt non-linear least squares method. The approximate values of the time derivatives were subsequently obtained at each sampled data point using the analytical derivative of Eqn. (28). For a related S-system identification problem [18] it was observed that that, typically, the accuracy of approximated derivative values is poor at the beginning and end of each signal and so, prior to the network inference procedure, we remove the first and last 4 data points from each signal for each experiment.

**6.3 DE settings**

Ten runs of the self-adaptive DE with the encoding described in Section 3.1 were performed for each reaction network in order to minimise the objective function in Eqn (18). DE, like other EAs, is a non-deterministic algorithm, and multiple runs are usually required to reliably assess its performance on a particular problem. Each run used a population size $P = 400$ and number of generations $G = 2000$. Runs were terminated early if the best value of the objective function obtained did not change for 500 generations. The value of $q_{max}$ (the maximum number of elementary reactions allowed in each trial network) was set to 10. Hence, in each case the DE searched for an optimal vector containing 50 integers, each of which can take a value between 0 and $S$, e.g. for the third reaction system $S = 10$ and so the entire search space contains $10^{50}$ reaction networks. All runs were performed using MATLAB v7.1 on a 1.8 GHz Windows PC with 2 Gb of RAM. The time taken for each run averaged at around 10 minutes.

## 6.4 Model selection methodology

Instead of choosing the "best" result (i.e. the lowest value of $SIC_D$) of each DE run as the solution to the network inference problem we instead performed a further step to choose a solution. This is because the trial network with the lowest $SIC_D$ is not necessarily the trial network that yields the lowest SIC when its ODEs are numerically integrated and compared directly with the measured concentration data. Errors introduced during the derivative approximation step can result in the DE search finding a trial network with the lowest $SIC_D$ but not necessarily the lowest SIC when numerically integrated. This is defined as $SIC_C$:

$$SIC_C = N_s \ln\left(\frac{SSE_C}{N_s}\right) + (p+\alpha)\ln(N_s) \qquad (29)$$

where $SSE_C$ is the sum of squared errors between the measured concentration data and the numerically integrated prediction of the ODEs corresponding to a trial network, all other terms are identical to those in the DE objective function defined in Eqn. (18). Hence, at the end of each DE run the ODES corresponding to each trial network in the entire final population were numerically solved and the predicted concentration values evaluated against the experimental data in terms of the $SIC_C$. The network with the lowest $SIC_C$ was considered to be the solution of the run, although in practice it would be desirable to consider several of the best performing networks from each run. These could be treated as a working portfolio of plausible reaction networks that could then be examined for chemical plausibility or as the starting point for a set of further experiments to discriminate between network hypotheses.

## 6.5 Results

The results of running the DE ten times for each of the hypothetical reaction networks are described below:

Reaction network 1

Each of the DE runs converged to the same value of $SIC_D$ = -696.29 corresponding to the reaction network described by Eqn. (30) with the following estimated rate coefficients $\hat{k}_1$ = 0.090, $\hat{k}_2$ = 0.130, $\hat{k}_3$ = 0.130, $\hat{k}_4$ = 0.304 and $\hat{k}_5$ = 0.073. Note that this network, although close to the true solution, is incorrect. It actually comprises the true network plus an additional fifth reaction (underlined in Eqn. (30)) which is stoichiometrically identical to the second reaction but gives rise to different kinetic terms in the corresponding ODEs. This is not a deficiency of the DE search process however, because the correct network topology actually yields a higher (i.e. worse) value of $SIC_D$ = -674.650. It seems likely that errors introduced by the derivative approximation process have led to overfitting whereby a network with an additional incorrect reaction fits the concentration derivative data better than the correct network.

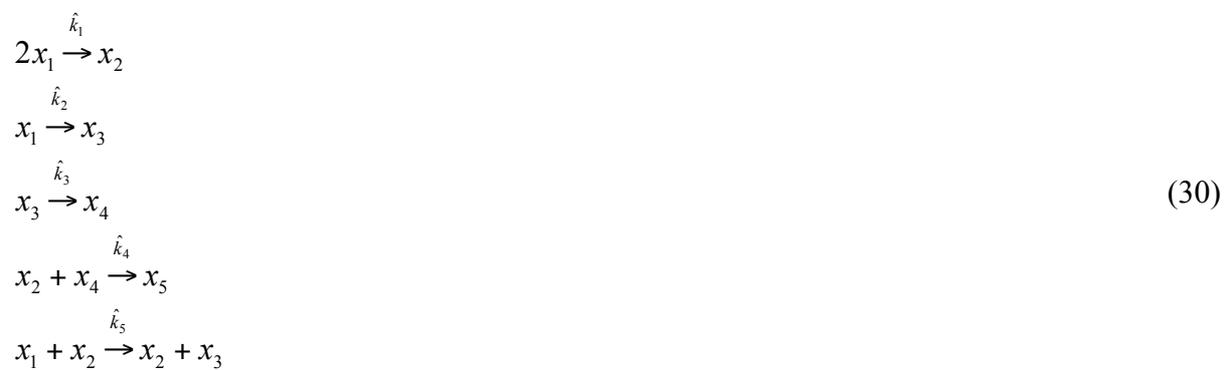

$$\begin{aligned} 2x_1 &\xrightarrow{\hat{k}_1} x_2 \\ x_1 &\xrightarrow{\hat{k}_2} x_3 \\ x_3 &\xrightarrow{\hat{k}_3} x_4 \\ x_2 + x_4 &\xrightarrow{\hat{k}_4} x_5 \\ \underline{x_1 + x_2} &\underline{\xrightarrow{\hat{k}_5} x_2 + x_3} \end{aligned} \qquad (30)$$

However, the application of the model selection methodology outlined in the previous section yielded the correct reaction network topology for each of the DE runs – showing that the correct solution, whilst not having the lowest $SIC_D$, was in each case present in the final population of each run and has an $SIC_C$ better than that obtained for the incorrect solution above. The estimated rate coefficients for the correct topology are $\hat{k}_1$ = 0.090, $\hat{k}_2$ = 0.191, $\hat{k}_3$ = 0.132 and $\hat{k}_4$ = 0.307 and are close to the true rate coefficients $k_1$ = 0.1, $k_2$ = 0.2, $k_3$ = 0.13 and $k_4 = 0.3$. Hence, it can be concluded that the true reaction network topology has been recovered from the data with satisfactory estimates of the rate coefficients.

Reaction network 2

The ten DE runs converged on solutions with $SIC_D$ values in the range -768.30 to – 778.79. Once again, in each case the network with the lowest $SIC_D$ of the run did not correspond to the true reaction network. Typically, the solutions found tend to indicate overfitting in that they comprise the correct network with incorrect additional reactions. For example, one such incorrect network is shown in Eqn. (31) with estimated rate coefficients $\hat{k}_1 = 0.2$, $\hat{k}_2 = 0.099$, $\hat{k}_3 = 0.153$, $\hat{k}_4 = 0.053$ and $\hat{k}_5 = 0.009$. Note that the rate coefficients for the correct reactions are close to the true coefficients and that the identified rate coefficient for the incorrect reaction is small in comparison to those for the correct reactions.

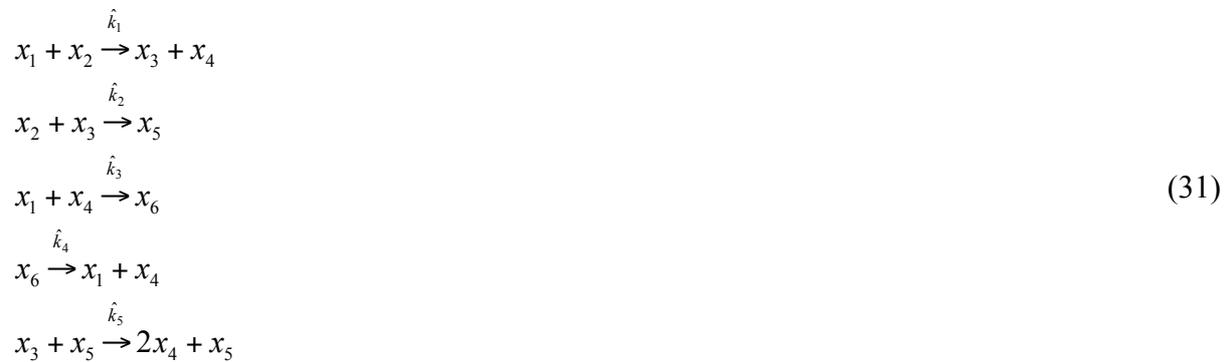

$$\begin{aligned} x_1 + x_2 &\xrightarrow{\hat{k}_1} x_3 + x_4 \\ x_2 + x_3 &\xrightarrow{\hat{k}_2} x_5 \\ x_1 + x_4 &\xrightarrow{\hat{k}_3} x_6 \\ x_6 &\xrightarrow{\hat{k}_4} x_1 + x_4 \\ x_3 + x_5 &\xrightarrow{\hat{k}_5} 2x_4 + x_5 \end{aligned} \qquad (31)$$

Again, the application of the model selection step on the final population of each DE run yielded, in each case, the correct reaction network topology with estimated rate coefficients $\hat{k}_1 = 0.200$, $\hat{k}_2 = 0.099$, $\hat{k}_3 = 0.155$ and $\hat{k}_4 = 0.058$. These are close to the true rate coefficients $k_1 = 0.2$, $k_2 = 0.1$, $k_3 = 0.15$ and $k_4 = 0.05$.

Reaction network 3

Unlike the previous hypothetical reaction networks, only six of the ten runs correctly identified the correct network topology after the $SIC_C$ model selection step. This is in contrast with the previous examples in which, in every case, the model selection step yielded the true network topology. This is not too surprising because this network was purposely designed to be a challenging problem in that it contains a "backbone" of unimolecular reactions in series, all with rate coefficients that are similar in magnitude. This means that many of the concentration time series profiles are highly correlated. Typically, the incorrectly identified networks are very similar to the true network in structure. For example, the incorrect network from run 6 is shown in Eqn. (32) and has estimated rate coefficients $\hat{k}_1 = 0.339$, $\hat{k}_2 = 0.249$, $\hat{k}_3 = 0.267$, $\hat{k}_4 = 0.409$, $\hat{k}_5 = 0.299$ and $\hat{k}_6 = 0.033$. The majority of the true network has been recovered correctly; the only incorrect reaction (underlined) contains the species $x_5$ as the reactant instead of the correct reactant $x_4$. Note that this incorrect reaction network satisfies the mass conservation constraint $\mathbf{AN} = 0$, as did the incorrect networks found in the other DE runs.

$$
\begin{aligned}
& x_1 + x_2 \xrightarrow{\hat{k}_1} x_3 \\
& x_3 \xrightarrow{\hat{k}_2} x_4 \xrightarrow{\hat{k}_3} x_5 \xrightarrow{\hat{k}_4} x_6 \\
& \underline{x_5 \xrightarrow{\hat{k}_5} x_7 + x_8} \\
& x_5 \xrightarrow{\hat{k}_6} x_9 + x_{10}
\end{aligned}
\qquad (32)
$$

In the six of the ten runs that yielded the correct network topology the estimated rate coefficients were $\hat{k}_1 = 0.338$, $\hat{k}_2 = 0.249$, $\hat{k}_3 = 0.300$, $\hat{k}_4 = 0.403$, $\hat{k}_5 = 0.299$ and $\hat{k}_6 = 0.100$. Again, these are very close to the true rate coefficients $k_1 = 0.35$, $k_2 = 0.25$, $k_3 = 0.3$, $k_4 = 0.4$, $k_5 = 0.3$ and $k_6 = 0.1$.

A further set of DE runs were performed that did not make use of the molecular weight information in the **A** matrix. In this case, only one of the ten runs yielded the correct network after the model selection step. Of the incorrect networks found in the other nine runs, each was tested and found to violate the **AN** = 0 conservation constraint. This emphasises the network search procedure is enhanced by the incorporation of prior chemical knowledge, especially on "difficult" reaction networks.

# 7 Conclusions

Reverse engineering the topology and dynamic behaviour of chemical and biological networks from noisy experimental time series data is a hard "inverse" problem. In this paper an evolutionary computational method for the identification of chemical reaction networks and the associated rate coefficients has been presented. Three case studies using data from simulated batch reactor experiments were used to demonstrate the technique and it was shown that it is possible to correctly identify the underlying chemical reaction networks and hence their stoichiometries and ODE models.

Advantages of our approach are: (a) it is almost entirely automated in that only a few parameters such as the DE population size $P$, number of generations $G$ and maximum network size $q_{max}$ need to be specified by the user (b) it can be applied to non-equilibrium chemical reaction systems, e.g. batch and semi-batch reactor experiments (c) in contrast to TFA methods, it is not necessary to separately determine either the number of independent reactions or reaction stoichiometries (d) the topology of the elementary reactions and the corresponding rate coefficients are estimated simultaneously (e) a priori chemical information in the form of reaction invariant constraints (e.g. molecular weights) can, when available, be incorporated into the network search procedure in a natural manner (f) the use of self-adaptive DE to search the space of elementary reaction networks provides a straightforward, robust optimisation method for the estimation of chemical reaction topologies and (g) the use of a population based search makes the subsequent use of a model selection procedure simple to apply.

Currently there are limitations to the method, principally: (a) it is assumed that concentration measurements are available for all of the species. When this is not true, however, it may be possible to employ existing TFA related methods [2] for estimating concentration data for unmeasured species (b) most chemical and biological networks tend to be sparsely connected and to promote parsimonious solutions a penalty function (soft constraint) approach was used in this work. It is likely, however, that the use of a multi-objective EA approach (e.g. see [26]) should improve the method by promoting diversity of network complexity in the population (c) the current implementation has only been tested on simulated reaction networks comprising elementary reactions and it is as yet unclear how method performs – or could be modified to perform - when presented with observed data from networks containing "complex" reactions, i.e. reactions comprising one or more elementary reactions where the intermediates cannot be measured and/or exist only fleetingly as transitional states.

Whilst the advantages of the incorporation of a priori chemical information have been demonstrated, it would be worthwhile considering the incorporation of additional system constraints in order to improve the robustness and reliability of the approach. Energy constraints, as well as those based on mass, could be used to enhance the network search process. For instance, a reaction – at constant temperature and pressure - can only occur if the Gibbs free energy of the reactants exceeds the Gibbs free energy of the products [27].

Future work will attempt to address the points discussed above in addition to testing the method on real experimental data from both "simple" reaction systems and from "complex" oscillating reaction networks such as that described in [28].

Finally we note that the success of the current approach is dependent on the ability to satisfactorily estimate first derivative values from sufficiently dynamically rich time series concentration data. The method used here (rational polynomials) is fairly simple but for real data more robust methods such as that recently described in [29] may be more appropriate.

**Acknowledgement**

This work was supported by the UK Engineering and Physical Research Council (EPSRC).